\documentclass[]{IEEEtran}
\IEEEoverridecommandlockouts
\usepackage{cite}
\usepackage{amsmath,amssymb,amsfonts}
\usepackage{graphicx}
\usepackage{textcomp}
\usepackage{amsmath}
\usepackage{booktabs} 
\usepackage{multirow}
\usepackage{hyperref}       
\usepackage{url}            
\usepackage{booktabs}       
\usepackage{amsfonts}       
\usepackage{nicefrac}       
\usepackage{microtype}      
\usepackage{xcolor}         
\usepackage{algorithm}
\usepackage{algpseudocode}
\usepackage{amsmath}
\usepackage{tikz}
\usetikzlibrary{quantikz2}
\usepackage{float}
\usepackage{rotating}
\usepackage{adjustbox}
\usepackage{setspace}

\usepackage{orcidlink}
\usepackage{comment}
\usepackage{amssymb}
\usepackage{braket}
\usepackage{xcolor}
\usepackage{stfloats}
\def\BibTeX{{\rm B\kern-.05em{\sc i\kern-.025em b}\kern-.08em
    T\kern-.1667em\lower.7ex\hbox{E}\kern-.125emX}}

\usepackage{enumitem}
\setlist[itemize]{leftmargin=*}%
\setlist[enumerate]{leftmargin=*}%
\usepackage[compact]{titlesec}
\usepackage{stfloats}

\titlespacing\section{0pt}{0.2\baselineskip}{0.12\baselineskip}
\titlespacing\subsection{0pt}{0.15\baselineskip}{0.08\baselineskip}
\titlespacing\subsubsection{0pt}{0.1\baselineskip}{0.08\baselineskip}

\begin{document}
\bstctlcite{bstctl:etal, bstctl:nodash, bstctl:simpurl}

\title{
Enhancing Blood Cells Classification using Hybrid Quantum Neural Networks
\vspace{-5pt}
}

\author{\IEEEauthorblockN{Guilherme Cruz\IEEEauthorrefmark{1}\orcidlink{0009-0006-9941-2022}, Nouhaila Innan\IEEEauthorrefmark{2}\IEEEauthorrefmark{3}\orcidlink{0000-0002-1014-3457}, Alberto Marchisio\IEEEauthorrefmark{2}\IEEEauthorrefmark{3}\orcidlink{0000-0002-0689-4776}, Gabriel Falcao\IEEEauthorrefmark{1}\IEEEauthorrefmark{4}\orcidlink{0000-0001-9805-6747},
Muhammad Shafique\IEEEauthorrefmark{2}\IEEEauthorrefmark{3}\orcidlink{0000-0002-2607-8135}}

\IEEEauthorblockA{\IEEEauthorrefmark{1} \normalsize Instituto de Telecomunicações, University of Coimbra, Portugal\\}
\IEEEauthorblockA{\IEEEauthorrefmark{2} \normalsize eBrain Lab, Division of Engineering, New York University Abu Dhabi, PO Box 129188, Abu Dhabi, UAE\\}
\IEEEauthorblockA{\IEEEauthorrefmark{3} \normalsize Center for Quantum and Topological Systems, NYUAD Research
Institute, New York University Abu Dhabi, UAE\\}
\IEEEauthorblockA{\IEEEauthorrefmark{4} \normalsize Science Division, New York University Abu Dhabi, UAE\\
Emails: gcruz@student.uc.pt, \{nouhaila.innan, alberto.marchisio, gabriel.falcao, muhammad.shafique\}@nyu.edu}
\vspace{-30pt}
}

\maketitle
\thispagestyle{empty} 
\begin{spacing}{0.93}
\begin{abstract}
Accurate classification of microscopic blood cells is still a critical task in medical image analysis, where subtle variations and limited data can challenge conventional deep learning models. As such, we investigate in this work the potential of Hybrid Quantum-Classical Neural Networks (HQNNs) to enhance feature representation and improve classification performance in this domain. 
We propose a modular architecture combining a pre-trained ResNet-50 backbone with a low-dimensional latent bottleneck and a variational quantum circuit, enabling a direct comparison between quantum-enhanced and purely classical transformation mechanisms.
To isolate the contribution of the quantum component, we evaluate three architectures: a HQNN model, a Classical Matched Model with an additional nonlinear transformation layer of comparable capacity, and a baseline model without an intermediate transformation stage.
Experiments conducted on two publicly available blood cell datasets, namely the Blood Cell Images dataset and the PBC dataset, demonstrate that HQNNs consistently achieve superior or more balanced performance across evaluation metrics. In the Blood Cell Images Dataset, the proposed approach improves macro F1-score by up to 3.7\% compared to classical baselines, while improving the F1-score from 98.54\% to 98.69\% in the more challenging 8-class scenario with near-saturated performance. Additional evaluation on IBM quantum hardware shows that the model remains robust under noise, with only a modest performance degradation relative to simulated results.
These results indicate that quantum feature transformations can enhance discriminative representations, particularly in challenging classification scenarios, and highlight the practical potential of HQNN models for medical imaging tasks.
\vspace{-5pt}
\end{abstract}

\begin{IEEEkeywords}
Quantum Machine Learning, Blood Cells Classification, Hybrid Quantum Neural Networks
\end{IEEEkeywords}
\end{spacing}

\begin{spacing}{0.93}
\vspace{-2pt}
\section{Introduction}

\begin{figure}[t!]
    \centering
    \includegraphics[height=1\columnwidth, angle=-90]{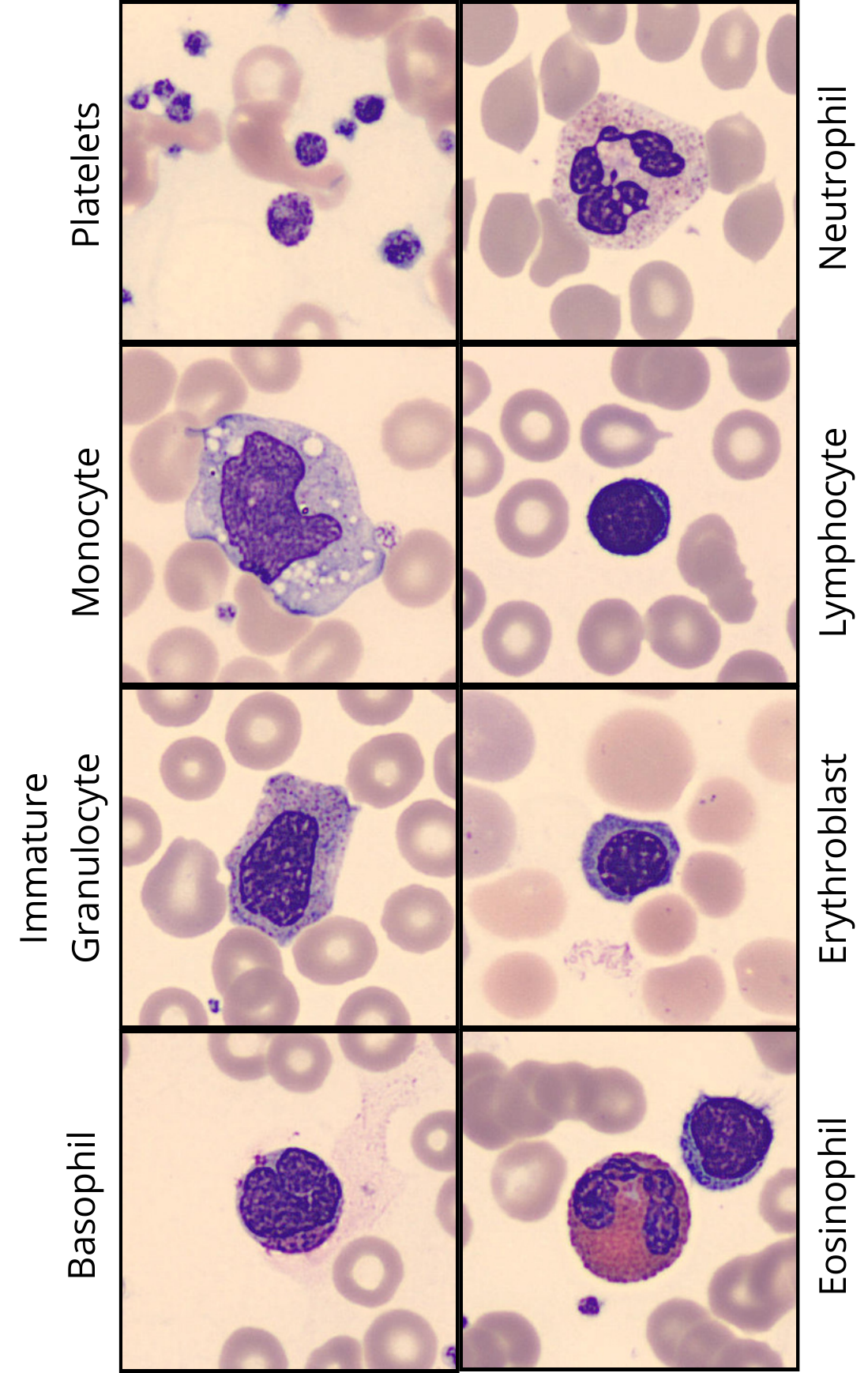}
    \vspace{-5pt}
    \caption{Representative examples of the eight classes in the PBC dataset.}
    \label{fig:8class_examples}
\end{figure}

\textbf{AI-based blood cell morphology analysis has emerged as an important support technology in hematology, where accurate differentiation of blood cells is essential for disease screening, diagnostic assessment, and therapy monitoring} \cite{xing2023artificial,fan2024microscope}. By reducing inter-observer variability and subjective interpretation errors while improving abnormal-cell detection, classification accuracy, and review efficiency, AI-assisted systems can strengthen blood-cell analysis workflows, although expert verification still remains necessary for challenging cases \cite{xing2023artificial,fan2024microscope}.
 Given the sheer volume of these tests performed daily, automating blood cell classification has become a major priority for clinical pipelines. Even incremental improvements in automated classification can significantly reduce clinician fatigue and lead to faster, more accurate patient diagnoses. \textbf{Historically, this analysis has been a manual, time-intensive process prone to inter-observer variability. While convolutional neural networks (CNNs) have largely mitigated these issues by learning hierarchical features directly from image data, current deep learning models still face distinct limitations} \cite{CWBCMLDLModels,girdhar2022wbc,bayat2022multiattention,islam2024explainable,karaddi2026custom,acevedo2019peripheral}. They frequently struggle with fine-grained classification tasks involving subtle morphological differences or overlapping visual traits, such as distinguishing between monocytes and neutrophils. As shown in Fig.~\ref{fig:8class_examples}, high intra-class variability further complicates this, acting as a bottleneck for the reliability of fully automated systems. 
Recently, researchers have turned to quantum computing as a potential solution for these edge cases \cite{WEI202342}. Hybrid Quantum-Classical Neural Networks (HQNNs) attempt to leverage the vast, high-dimensional space of quantum Hilbert spaces and the complex correlations afforded by quantum entanglement \cite{QCLearning,CCQClassifiers,VQAlgorithms,Liu_2021,HQCQ}. Recent healthcare-oriented HQNN studies have explored both broader biomedical learning settings and blood-image analysis, although direct multi-class blood cell classification with HQNN models remains scarce \cite{innan2024fedqnn,bano2026analyzingimagesbloodcells}. \textbf{This study isolates and investigates the specific role of quantum feature transformations in blood cell classification.} We introduce a comparative framework built around a modular architecture: a pre-trained ResNet-50 backbone coupled with a low-dimensional latent bottleneck. For the intermediate transformation stage, the architecture can alternate between a variational quantum circuit and a classical nonlinear mapping. \textbf{This setup guarantees a strictly controlled comparison between purely classical and hybrid approaches.} We evaluate these models across two public datasets with varying complexity (a 4-class and an 8-class task) and further assess the HQNN model's real-world robustness by executing it on IBM quantum hardware.

\textbf{The novel contributions of this work can be summarized as follows:}

\begin{itemize}
    \item \textbf{Controlled hybrid vs. classical comparison framework:} We propose a modular architecture that enables a strictly controlled comparison between HQNNs and purely classical models by maintaining identical backbones, bottleneck dimensions, and classifier heads, while only varying the intermediate transformation stage.

    \item \textbf{Fair classical baselines with matched capacity:} To ensure a rigorous evaluation, we introduce multiple classical baselines, including (i) a model without the quantum layer and (ii) an enhanced classical model with an additional nonlinear layer of comparable parameter count to the quantum circuit, ensuring a fair capacity-matched comparison.

    \item \textbf{Real quantum hardware validation:} Beyond simulation, we demonstrate the practical feasibility of the proposed HQNN model by executing inference on IBM quantum hardware, analyzing robustness under realistic noise and hardware constraints.

    \item \textbf{Empirical analysis of quantum advantage in medical imaging:} Through extensive experiments, we isolate and quantify the contribution of the quantum feature transformation, showing that performance gains are more pronounced in challenging classification scenarios characterized by high intra-class variability and subtle inter-class differences.
\end{itemize}

\section{Background and Related Work}

\subsection{Blood Cells Classification}

Automating blood cell classification is critical for overcoming the subjectivity, inconsistency, and inefficiency of manual microscopic review. Historically, diagnostic pipelines relied heavily on handcrafted techniques, chaining together image segmentation, morphological feature extraction, and traditional machine learning classifiers~\cite{saraswat2014automated, rezatofighi2011automatic}. While these early methods functioned well in highly controlled laboratory environments, their reliability degraded sharply when exposed to real-world clinical variables like irregular staining, fluctuating illumination, and natural morphological diversity within cell classes~\cite{sarrafzadeh2017best}.

The shift toward deep learning, specifically convolutional neural networks (CNNs), fundamentally altered this landscape by allowing models to learn robust, hierarchical features directly from raw image pixels. Current standard practice relies heavily on transfer learning. By fine-tuning established architectures like ResNet, VGG, or Inception that were pre-trained on massive generalized datasets, researchers have achieved highly accurate diagnostic baseline models~\cite{habibzadeh2018automatic}. These CNN-based frameworks perform exceptionally well on standard 4-class problems (eosinophils, lymphocytes, monocytes, neutrophils) and have scaled successfully to more demanding 8-class taxonomies requiring finer morphological distinctions.

More recently, the integration of attention mechanisms, aggressive data augmentation, and targeted regularization has pushed model accuracy even further. Attention modules allow networks to dynamically isolate and prioritize diagnostically critical regions of the cell, pushing performance on standard public benchmarks to near-saturation levels~\cite{chen2022accurate}. Naturally, the growing availability of large, expertly annotated medical datasets has been instrumental in training these increasingly sophisticated architectures~\cite{kouzehkanan2022large}.

Nevertheless, the current deep learning paradigm faces distinct bottlenecks. Fine-grained classification remains a persistent hurdle; when morphological boundaries between distinct cell types blur and intra-class variance is high, classical CNN feature maps often overlap, leading to misclassification of visually similar cells. Furthermore, these architectures are notoriously data-hungry, demanding vast annotated repositories that are expensive and difficult to curate in specialized medical fields. Finally, standard neural networks are fundamentally bound by the mathematical limits of classical vector spaces. This structural constraint restricts their ability to map the highly complex, non-linear correlations hidden within high-dimensional biomedical imagery.

Recognizing these limitations highlights the need for novel computational frameworks. \textit{As classical models approach a performance ceiling, it becomes essential to explore alternative paradigms capable of expanding a model's representational capacity and resolving ambiguous morphological features without demanding an exponential increase in training data.}

\subsection{Hybrid Quantum Neural Networks}

Merging quantum computing principles with classical deep learning pipelines has introduced a new paradigm: Hybrid Quantum-Classical Neural Networks (HQNNs)~\cite{arthur2022hybrid}. These architectures generally pair classical feature extractors, such as CNN backbones, with variational quantum circuits (VQCs) that serve as intermediate transformation layers~\cite{VQAlgorithms}. The core rationale is that mapping data into the high-dimensional space of HQNNs, coupled with the expressive dynamics of quantum entanglement, might reveal complex feature correlations that classical architectures struggle to model.

For image classification, these pipelines generally follow a specific sequence: the network compresses classical features into a low-dimensional latent representation, which is then embedded into a quantum state~\cite{senokosov2024quantum}. A parameterized quantum circuit then processes this state, and the resulting measurements pass directly into a classical classifier. Functionally, this allows the quantum layer to act as a highly non-linear feature transformer operating within a fundamentally distinct computational environment. Recent quantum image classification frameworks have also explored replacing or modifying components of classical CNN pipelines with variational quantum circuits, reporting improved accuracy and F1-score while reducing parameter complexity in some settings~\cite{NICFrameworkVQA, kashif2025computational}.

\begin{figure*}[t!]
    \centering
    \includegraphics[width=.9\textwidth]{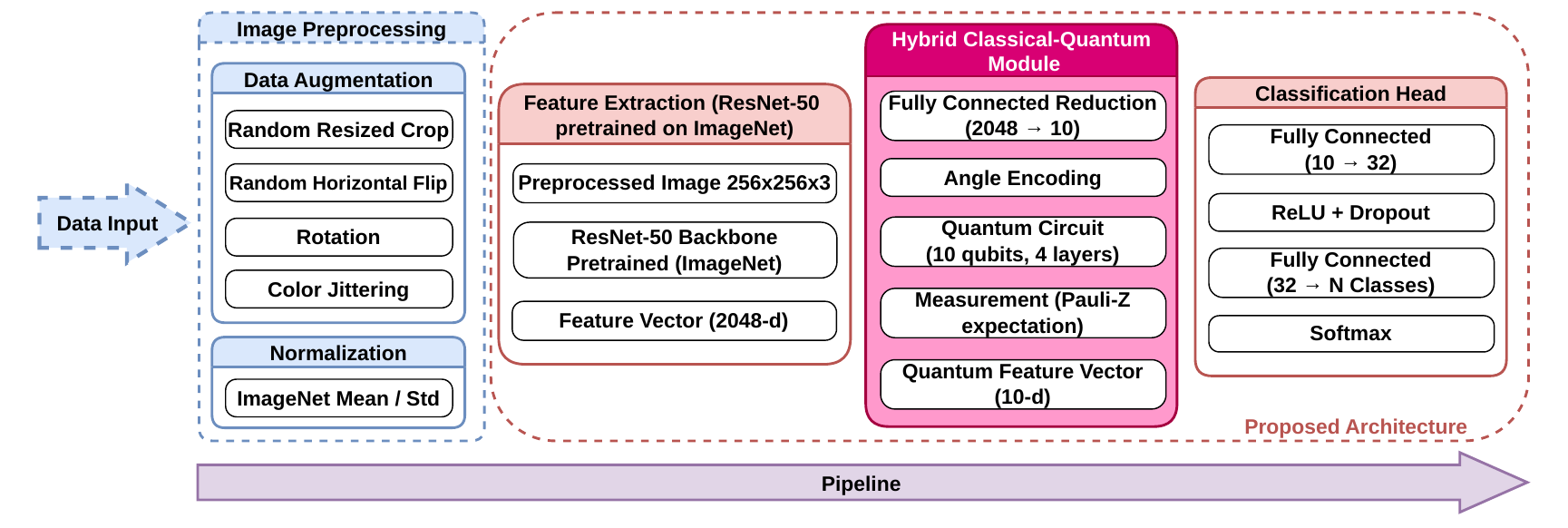}
    \vspace{-8pt}
    \caption{Overview of the proposed blood cell classification pipeline. The framework processes both the Blood Cell Images and PBC datasets through a common preprocessing and augmentation stage, followed by feature extraction using a pre-trained ResNet-50 backbone. The resulting 2048-dimensional feature vector is reduced to a 10-dimensional latent representation and passed to an HQNN composed of angle encoding, a variational quantum circuit with 10 qubits and 4 layers, and Pauli-Z measurement. The resulting quantum feature vector is then fed to a lightweight classification head for final blood cell type prediction.}
    \label{fig:architecture}
\end{figure*}

While early HQNN applications in medical imaging suggest advantages in representation learning, empirical evidence remains mixed, with modest improvements often fluctuating based on dataset and model design. A major hurdle is isolating the quantum contribution; gains are frequently attributed to classical architectural shifts or parameter counts rather than intrinsic quantum advantages. Validating these systems requires rigorous, controlled experiments against matched classical counterparts.

Beyond theoretical evaluation, practical hardware constraints, such as prohibitive quantum noise, restricted qubit counts, and the steep computational overhead of simulating quantum states, continue to limit scalability. Since most contemporary research relies on software simulators instead of physical quantum devices, the real-world viability of these pipelines remains an open question.

Ultimately, while HQNNs offer a compelling theoretical avenue for advancing machine learning, their tangible utility in medical image classification, particularly for blood cell analysis, is largely unexplored. \textit{By systematically isolating the quantum layer and measuring its impact against strict classical baselines, this work aims to clarify the genuine diagnostic value of quantum feature transformations}.

\section{Methodology}
\subsection{System Overview}

The proposed model follows a modular processing pipeline consisting of three main stages: feature extraction, dimensionality reduction, and classification. First, high-level visual features are extracted from blood cell images using a pre-trained ResNet-50 model. The high-dimensional feature vectors produced by the backbone network are then projected into a lower-dimensional latent representation through a classical feedforward layer performing a linear transformation of the ResNet output. The resulting latent representation is subsequently processed using one of two transformation mechanisms, depending on the evaluated architecture: either a variational quantum circuit or a classical transformation designed to provide a comparable representational capacity. The resulting transformed features are then passed to a lightweight classifier responsible for predicting the corresponding blood cell class.

To analyze the contribution of the quantum component, three different architectures are evaluated. The first architecture corresponds to an HQNN model, in which the reduced feature representation is processed through a variational quantum circuit implemented using PennyLane. The second architecture replaces the quantum circuit with an additional classical layer containing a comparable number of parameters and occupying the same functional position in the pipeline, enabling a direct comparison between quantum and classical representations. The third architecture removes this intermediate transformation stage entirely, resulting in a purely classical baseline model. This experimental design allows us to determine whether potential improvements arise from the quantum transformation itself or from the increased representational capacity introduced by an additional transformation layer. The complete system architecture and the interaction between its components are illustrated in Fig.~\ref{fig:architecture}.

The performance of the proposed framework is evaluated using two datasets of microscopic blood cell images containing four and eight classification categories, respectively. All architectures are trained under identical conditions to ensure a fair comparison across models. The following subsections describe the HQNN architecture, the design of the quantum layer, and the training procedure in detail.

\subsection{Data Preprocessing and Augmentation}
To ensure a fair comparison across all model variants, the same preprocessing pipeline was applied to the HQNN model and to the classical baselines in both the Blood Cell Images and PBC datasets experiments. All input images were resized to a fixed resolution before being passed to the network. During training, data augmentation was used to improve robustness to small changes in cell position, orientation, and visual appearance, which are common in microscopic blood cell images. The augmentation pipeline included random cropping, horizontal flipping, small rotations, and color based perturbations. After these transformations, the images were converted into tensor format and normalized using the standard statistics commonly adopted for ImageNet pretrained models. This same preprocessing strategy was used consistently for all architectures and dataset settings.

For evaluation, no random augmentation was applied. The validation and test images were only resized, converted into tensors, and normalized with the same statistics used during training. This choice ensures a deterministic evaluation protocol and makes the comparison between models more reliable.

\subsection{Feature Extraction Using ResNet-50}
A ResNet-50 network was used as the backbone feature extractor in all model variants. The same backbone was kept for the HQNN architecture and for both classical baselines so that the comparison would focus on the effect of the intermediate transformation stage rather than on differences in the convolutional encoder. In all cases, the model was initialized with ImageNet pretrained weights, and its original final classification layer was removed. This allows the network to act only as a visual feature extractor and to produce a compact representation of each input image before the subsequent processing stages. In the implemented model, the backbone outputs a feature vector with 2048 dimensions, which is then passed to the reduction layer described in the next subsection.

The use of the same ResNet-50 backbone across all experiments also helps maintain a controlled setting for the comparison between quantum and classical models. At the beginning of training, the convolutional backbone is loaded with pretrained parameters, while the newly added layers are kept trainable. During fine tuning, the backbone is unfrozen, although the earliest layers remain fixed. More specifically, the initial convolution and batch normalization layers are kept frozen, while the remaining backbone layers are updated together with the new task specific layers. This strategy preserves low level visual filters while allowing the deeper representations to adapt to the blood cell classification task.

\subsection{Feature Reduction and Latent Representation}

After feature extraction, the 2048 dimensional output of the ResNet-50 backbone is mapped to a much smaller latent space through a fully connected reduction layer. This step creates a compact representation that can be processed consistently by all model variants. In the implemented architecture, the bottleneck dimension is set to 10, which also matches the number of qubits used in the quantum branch. The reduction stage therefore acts as the interface between the convolutional backbone and the later transformation modules.
The reduced features are passed through a hyperbolic tangent activation before the next stage. In the HQNN model, this keeps the latent values in a bounded interval, which is convenient for angle based quantum encoding. The same bounded representation is also kept in the classical variants so that the comparison remains consistent across architectures. As a result, the three models share the same feature extractor, the same reduction stage, and the same latent dimensionality, differing only in the transformation applied after the bottleneck.
\begin{equation}
\mathbf{z} = \tanh\!\left(\mathbf{W}_{r}\mathbf{f} + \mathbf{b}_{r}\right),
\qquad
\mathbf{f} \in \mathbb{R}^{2048},
\quad
\mathbf{z} \in \mathbb{R}^{10}.
\end{equation}

\subsection{Variational Quantum Circuit Design}
The quantum branch is implemented as a variational quantum circuit built with PennyLane. The circuit uses 10 qubits and 4 trainable variational layers, and it is executed on the \texttt{lightning.gpu} simulator. In the present implementation, the selected encoding strategy is angle embedding, where each component of the reduced latent vector is injected into the circuit through rotations around the Y axis. This choice allows the classical bottleneck features to be mapped directly into the quantum state in a simple and computationally efficient way.

After the encoding step, the circuit applies trainable rotation gates followed by entangling operations. More specifically, each variational layer contains one trainable $R_Y$ rotation per qubit and a ring structured pattern of controlled NOT gates to introduce correlations between neighboring qubits. At the output, one expectation value of the Pauli Z observable is measured from each qubit, producing a 10 dimensional quantum feature vector. These measured values are then passed to the final classifier. The trainable quantum parameters are initialized with small random values, and gradients are computed through the adjoint differentiation method.

\begin{figure}[t!]
\centering
\resizebox{\columnwidth}{!}{%
\begin{quantikz}[row sep=0.35cm, column sep=0.28cm]
\lstick{\(q_0\)}
  & \gate{R_Y(z_0)}
  & \gate{R_Y(\theta_{1,0})} & \ctrl{1} & \qw      & \targ{}    
  & \gate{R_Y(\theta_{2,0})} & \ctrl{1} & \qw      & \targ{}    
  & \gate{R_Y(\theta_{3,0})} & \ctrl{1} & \qw      & \targ{}    
  & \gate{R_Y(\theta_{4,0})} & \ctrl{1} & \qw      & \targ{}    
  & \meter{} & \rstick{\(\langle Z_0\rangle\)} \\
\lstick{\(q_1\)}
  & \gate{R_Y(z_1)}
  & \gate{R_Y(\theta_{1,1})} & \targ{}  & \ctrl{1} & \qw        
  & \gate{R_Y(\theta_{2,1})} & \targ{}  & \ctrl{1} & \qw        
  & \gate{R_Y(\theta_{3,1})} & \targ{}  & \ctrl{1} & \qw        
  & \gate{R_Y(\theta_{4,1})} & \targ{}  & \ctrl{1} & \qw        
  & \meter{} & \rstick{\(\langle Z_1\rangle\)} \\
\lstick{\(q_2\)}
  & \gate{R_Y(z_2)}
  & \gate{R_Y(\theta_{1,2})} & \qw      & \targ{}  & \ctrl{-2}  
  & \gate{R_Y(\theta_{2,2})} & \qw      & \targ{}  & \ctrl{-2}  
  & \gate{R_Y(\theta_{3,2})} & \qw      & \targ{}  & \ctrl{-2}  
  & \gate{R_Y(\theta_{4,2})} & \qw      & \targ{}  & \ctrl{-2}  
  & \meter{} & \rstick{\(\langle Z_2\rangle\)}
\end{quantikz}
}
\caption{Illustrative schematic of the variational quantum circuit used in the proposed hybrid model. Three qubits are shown to communicate the ring CNOT entanglement pattern: each qubit controls the next (\(q_i \rightarrow q_{i+1}\)), with the final qubit closing the ring back to the first (\(q_2 \rightarrow q_0\), and equivalently \(q_9 \rightarrow q_0\) in the full 10-qubit circuit). The full model uses 10 qubits and four variational layers, where each layer applies trainable \(R_Y(\theta_{\ell,i})\) rotations followed by ring entanglement.}
\label{fig:qvc_main}
\end{figure}
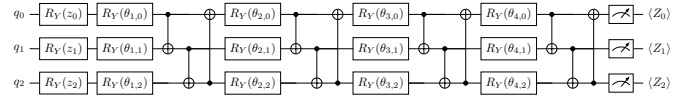

The corresponding circuit topology is illustrated in Fig.~\ref{fig:qvc_main} using a three-qubit schematic for readability.
\begin{equation}
U_{\mathrm{enc}}(\mathbf{z})
=
\bigotimes_{i=1}^{n_q} R_Y(z_i),
\qquad n_q = 10.
\end{equation}
\begin{equation}
U(\mathbf{z}, \Theta)
=
\left(
\prod_{\ell=1}^{L}
U_{\mathrm{var}}^{(\ell)}(\boldsymbol{\theta}^{(\ell)})
\right)
U_{\mathrm{enc}}(\mathbf{z}),
\qquad L = 4.
\end{equation}
\begin{equation}
q_i
=
\left\langle 0^{\otimes n_q} \right|
U^{\dagger}(\mathbf{z}, \Theta)\,
Z_i\,
U(\mathbf{z}, \Theta)
\left| 0^{\otimes n_q} \right\rangle,
\qquad i = 1, \dots, n_q.
\end{equation}
\subsection{Classical and Hybrid Model Architectures}
To isolate the effect of the quantum component, three architectures were evaluated under the same general pipeline. The first one is the HQNN model. In this case, the reduced latent representation is sent to the variational quantum circuit, and the measured quantum outputs are then used for classification. This is the main architecture proposed in this work and the one used to study the contribution of quantum feature transformation to blood cell recognition.

The second model is the Classical Matched Model, a classical comparison architecture designed to mirror the hybrid pipeline as closely as possible. Instead of the quantum circuit, it introduces an additional classical block after the bottleneck. This block is implemented as a linear transformation followed by a hyperbolic tangent activation, and it preserves the same latent dimensionality. Its role is to provide an extra nonlinear transformation stage with a similar functional position in the pipeline, allowing a more controlled comparison between quantum and classical intermediate processing.

The third model is a simpler baseline that removes the intermediate transformation block entirely. In this version, the reduced latent representation is passed directly to the classifier head. This baseline makes it possible to distinguish between gains that may come from the quantum circuit itself and gains that may simply come from adding another transformation stage after the bottleneck. Because all three models share the same preprocessing, backbone, bottleneck size, and classifier structure, the comparison remains centered on the effect of the intermediate representation learning stage.

\begin{table}[t!]
\centering
\caption{Detailed parameter breakdown per component across the evaluated architectures.}
\label{tab:model_param_breakdown}
\resizebox{\columnwidth}{!}{%
\begin{tabular}{lccc}
\toprule
Component & Classical Baseline & Classical Matched Model & HQNN \\
\midrule
ResNet50 Backbone &  23{,}508{,}032 & 23{,}508{,}032 & 23{,}508{,}032 \\
fc\_reduce &  20{,}490 & 20{,}490 & 20{,}490 \\
classical\_block  & --- & 110 & --- \\
q\_layer.weights & --- & --- & 40 \\
head & 616 & 616 & 616 \\
\midrule
\textbf{Total} &  \textbf{23{,}529{,}138} & \textbf{23{,}529{,}248} & \textbf{23{,}529{,}178} \\
\bottomrule
\end{tabular}%
}
\end{table}

The parameter breakdown in Table~\ref{tab:model_param_breakdown} highlights that all three architectures share the same dominant components, namely the ResNet-50 backbone, the feature reduction layer, and the classification head. As a result, the total number of parameters remains nearly identical across models, with only minor differences introduced by the intermediate transformation stage. 

In particular, the Classical Matched Model adds 110 trainable parameters through an additional nonlinear transformation layer, while the HQNN introduces 40 trainable parameters within the variational quantum circuit. This controlled design is critical for isolating the effect of the intermediate transformation mechanism.
\begin{equation}
\mathbf{z}^{\prime}
=
\tanh\!\left(\mathbf{W}_{c}\mathbf{z} + \mathbf{b}_{c}\right).
\end{equation}
\subsection{Classification Head}
The final prediction stage is implemented with a lightweight fully connected classifier. After the intermediate transformation step, whether quantum, classical, or direct, the resulting 10 dimensional representation is passed to a linear layer with 32 hidden units, followed by a ReLU activation and dropout. A final linear layer then maps this representation to the target number of classes. The same classification head is used in all model variants, which helps maintain a fair comparison between them.

This design keeps the final decision stage intentionally simple. The main idea is that the backbone and the intermediate transformation module should perform most of the representation learning, while the classifier head remains small and task specific. The use of dropout also provides a regularizing effect, which is useful given the limited size and visual similarity often found in blood cell image datasets.
\begin{equation}
\hat{\mathbf{y}}
=
\mathbf{W}_{2}\,
\phi\!\left(\mathbf{W}_{1}\mathbf{h} + \mathbf{b}_{1}\right)
+
\mathbf{b}_{2}
\end{equation}
where $\mathbf{h}$ denotes the input to the classifier head, and $\phi(\cdot)$ represents the ReLU activation function.
\subsection{Training Strategy}
All models are trained through end to end fine tuning. At the beginning of training, the pretrained ResNet-50 backbone is unfrozen, but the earliest convolution and batch normalization layers remain fixed. This preserves low level visual filters while allowing the deeper layers and the newly introduced modules to adapt to the target task. For optimization, Adam is used with separate parameter groups. In the HQNN model, different learning rates are assigned to the backbone, the newly added classical layers, and the quantum layer. In the classical baselines, separate learning rates are used for the backbone and the task specific classical modules.

Training is further stabilized with a cosine annealing learning rate scheduler, mixed precision training, and gradient clipping. The loss function is focal loss with label smoothing, which helps the model focus more on difficult samples while reducing overconfidence. During training, validation macro F1-score is monitored at the end of each epoch, and early stopping is applied when no improvement is observed for a fixed number of epochs. The best performing checkpoint is then restored for the final evaluation. The code also fixes random seeds for Python, NumPy, and PyTorch to improve reproducibility across runs.

Let $\mathcal{L}_{\mathrm{CE}}^{(m)}$ denote the label smoothed cross entropy loss for sample $m$.

\begin{equation}
\mathcal{L}_{\mathrm{focal}}^{(m)}
=
\left(
1 - e^{-\mathcal{L}_{\mathrm{CE}}^{(m)}}
\right)^{\gamma}
\mathcal{L}_{\mathrm{CE}}^{(m)},
\end{equation}

\begin{equation}
\mathcal{L}
=
\frac{1}{N}
\sum_{m=1}^{N}
\mathcal{L}_{\mathrm{focal}}^{(m)}.
\end{equation}

\subsection{Evaluation Metrics}
Model performance is evaluated through both aggregate and class-wise measures. The implemented evaluation pipeline computes per class accuracy, sensitivity, precision, and F1-score using a one-versus-rest formulation, and then reports the macro-average across classes. This choice is appropriate for multiclass medical image classification because it gives equal importance to all classes, including those that may be more difficult or less frequent. In addition to these metrics, overall accuracy is also reported during training and evaluation for a more direct view of prediction quality.

To complement the threshold based measures, the evaluation also includes multiclass ROC AUC. This is computed from the softmax probabilities produced by the model and is reported in both macro and weighted forms. Confusion matrices are also generated to support the class-wise analysis of prediction errors. Together, these measures provide a broader view of model behavior and make it possible to compare not only overall performance but also class separability and error patterns across the different architectures.

\section{Results and Discussion}
\subsection{Experimental Setup}

The experimental evaluation was conducted using two publicly available blood cell image datasets obtained from Kaggle. These datasets were selected to assess the proposed models under different classification granularities, namely a 4-class and an 8-class setting.

The first dataset corresponds to the Blood Cell Images Dataset, a 4-class blood cell classification dataset derived from the BCCD dataset~\cite{bccd_dataset} and accessed through the Kaggle repository curated by Mooney~\cite{mooney_blood_cells}. It contains augmented microscopic blood cell images distributed across four categories: eosinophils, lymphocytes, monocytes, and neutrophils.

The second dataset corresponds to the Peripheral Blood Cell (PBC) dataset introduced by Acevedo et al.~\cite{acevedo2019peripheral}, which contains 17,092 images spanning eight distinct blood cell types. Following the official split (64\%/24\%/12\% for training, validation, and testing), the training subset is augmented to balance the class distribution, resulting in a training set of 32,776 images (4,097 per class), while the validation (4,098 images) and test (2,056 images) sets remain unchanged.

For the Blood Cell Images dataset, the original \texttt{TRAIN} folder contains 9,957 images and the \texttt{TEST} folder contains 2,487 images. As illustrated in Table~\ref{tab:dataset_splits}, the training set was further split into 85\% training and 15\% validation subsets ($\approx$ 8,463 and 1,494 images, respectively), while the original \texttt{TEST} folder was used as the held-out evaluation set.
Validation performance was used for model selection and early stopping, and final results were reported on the test sets for both datasets.

\begin{table}[t!]
\centering
\caption{Dataset split summary used in the experiments.}
\label{tab:dataset_splits}
\begin{adjustbox}{max width=\linewidth}
\footnotesize
\begin{tabular}{lcccc}
\toprule
Dataset & Classes & Train & Validation & Test \\
\midrule
Blood Cell Images & 4 & $\sim$8,463 & $\sim$1,494 & 2,487 \\
PBC dataset & 8 & 32,776 & 4,098 & 2,056 \\
\bottomrule
\end{tabular}%
\end{adjustbox}
\end{table}

Both datasets follow a directory-based structure compatible with PyTorch’s ImageFolder interface, where each subdirectory represents a class label. The training splits of both datasets are balanced, ensuring uniform class representation and reducing bias during model optimization.

The models are trained using the Adam optimizer over multiple epochs with a moderate batch size. Separate learning rates are assigned to different components of the architecture, with a lower learning rate used for the pre-trained ResNet-50 backbone and higher learning rates applied to the task-specific classification layers. In the HQNN models, the quantum layer is optimized independently using its own learning rate to account for its distinct training dynamics. Tables~\ref{tab:quantum_hyperparameters} and~\ref{tab:training_hyperparameters} report the hyperparameters used in the model configurations and training, respectively.

\begin{table}[t!]
\centering
\caption{Quantum-specific hyperparameters used in the HQNNs}
\label{tab:quantum_hyperparameters}
\begin{adjustbox}{max width=\linewidth}
\begin{tabular}{lccccc}
\toprule
Dataset & Qubits & Q-Layers & Encoding & Entanglement & LR$_{Q}$\\
\midrule
Blood Cell Images & 10 & 4 & Angle & Ring & $1\!\times\!10^{-4}$\\
PBC dataset & 10 & 4 & Angle & Ring & $1\!\times\!10^{-4}$\\
\bottomrule
\end{tabular}
\end{adjustbox}
\end{table}

\begin{table}[t!]
\centering
\caption{Training and optimization hyperparameters}
\label{tab:training_hyperparameters}
\begin{adjustbox}{max width=\linewidth}
\begin{tabular}{lccccc}
\toprule
Dataset & Batch & Epochs & Patience & LR$_{bb}$ & LR$_{head}$ \\
\midrule
Blood Cell Images & 16 & 70 & 15 & $1\!\times\!10^{-5}$ & $5\!\times\!10^{-5}$ \\
PBC dataset & 16 & 70 & 15 & $1\!\times\!10^{-5}$ & $5\!\times\!10^{-5}$ \\
\bottomrule
\end{tabular}
\end{adjustbox}
\end{table}

To improve robustness and emphasize hard-to-classify samples, a Focal Loss function with label smoothing is employed during training. Training stability and computational efficiency are further enhanced through mixed-precision computation and gradient scaling.

Model performance is evaluated using a comprehensive set of metrics, including overall accuracy, macro-averaged F1-score, precision and recall (sensitivity). In addition, multiclass ROC-AUC is computed using both macro and weighted averaging strategies. Per-class metrics and confusion matrices are also reported to provide a more detailed analysis of classification behavior across different blood cell types.

All experiments are implemented in PyTorch and executed on GPU-enabled hardware. The experimental platform consists of an NVIDIA GeForce RTX 5080 GPU with 16GB of VRAM and 128GB of RAM. For the HQNN models, the quantum component is implemented using the PennyLane framework with the \texttt{lightning.gpu} backend, allowing accelerated simulation of variational quantum circuits.

\subsection{Blood Cell Images Dataset Results}

\subsubsection{Convergence Analysis}

\begin{figure}[t!]
    \centering
    \includegraphics[width=\linewidth]{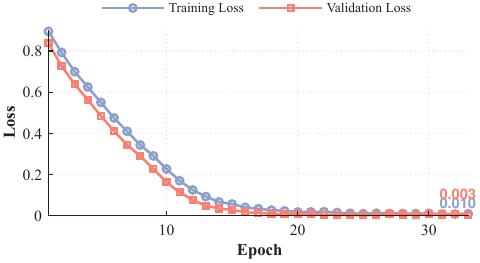}
    \vspace{-15pt}
    \caption{Training and validation loss curves of the HQNN model on the Blood Cell Images Dataset (4 Classes).}
    \label{fig:4class_training_loss}
\end{figure}

\begin{figure}[t!]
    \centering
    \includegraphics[width=\linewidth]{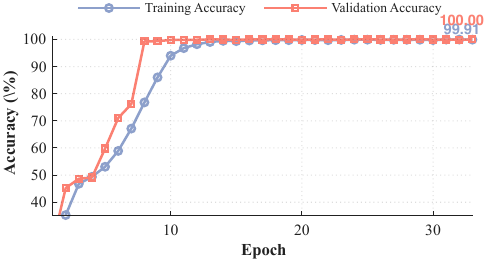}
    \vspace{-15pt}
    \caption{Training and validation accuracy curves of the HQNN model on the Blood Cell Images Dataset.}
    \label{fig:4class_training_accuracy}
\end{figure}

Figures~\ref{fig:4class_training_loss} and~\ref{fig:4class_training_accuracy} show the training and validation loss and accuracy evolution on the Blood Cell Images Dataset of the HQNN model, respectively. The loss decreases steadily during the early epochs and then stabilizes, indicating a stable optimization process. The smooth convergence profile suggests that the quantum transformation does not introduce training instability.

Overall, the HQNN model exhibits reliable convergence and reaches a low final loss, which is consistent with the improved macro-level classification results reported in the following subsection.


\subsubsection{Overall Performance}

The overall performance of the evaluated models on the Blood Cell Images Dataset, in terms of overall accuracy, macro-averaged precision, recall, and F1-score, together with macro and weighted ROC-AUC, is summarized in Table~\ref{tab:4class_overall_results}.

\begin{table}[t!]
\centering
\caption{Overall performance comparison on the Blood Cell Images Dataset}
\label{tab:4class_overall_results}
\resizebox{\columnwidth}{!}{%
\begin{tabular}{lccc}
\toprule
Metric & Classical Matched & Classical Baseline & HQNN \\
\midrule
Accuracy         & 0.9429 & 0.9375 & \textbf{0.9572} \\
Macro Precision  & 0.8975 & 0.9063 & \textbf{0.9176} \\
Macro Recall     & 0.8858 & 0.8749 & \textbf{0.9145} \\
Macro F1         & 0.8862 & 0.8784 & \textbf{0.9153} \\
ROC-AUC Macro    & 0.9745 & 0.9633 & \textbf{0.9597} \\
ROC-AUC Weighted & 0.9745 & 0.9633 & \textbf{0.9597} \\
\bottomrule
\end{tabular}%
}
\end{table}

All models achieved high performance, reflecting the relatively low complexity of the task and the balanced nature of the dataset. However, clear differences emerge when comparing the macro-level metrics.

\begin{figure}[t!]
    \centering
    \includegraphics[width=\linewidth]{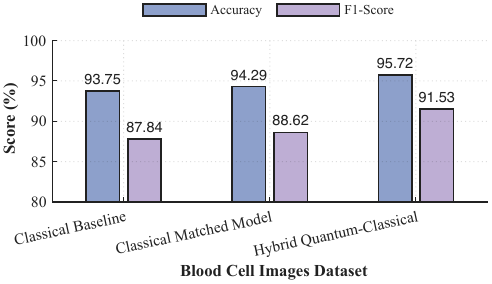}
    \vspace{-15pt}
    \caption{Comparison of overall accuracy and macro F1-score for the evaluated models on the Blood Cell Images Dataset.}
    \label{fig:4class_f1_acc}
\end{figure}

As shown in Fig.~\ref{fig:4class_f1_acc}, the HQNN model achieved the best overall performance, with a macro F1-score of 0.9153, outperforming both the Classical Matched model (0.8862) and the classical baseline (0.8784). This corresponds to an improvement of approximately 3.0\%–3.7\% in macro F1-score.

Similarly, the HQNN model achieved the highest macro precision (0.9176) and macro recall (0.9145), indicating improved consistency across all classes. In terms of overall accuracy, the HQNN model reached 0.9572, compared with 0.9429 for the Classical Matched model and 0.9375 for the classical baseline.

The ROC-AUC results show that all models achieve high values, indicating strong class separability overall. Although the HQNN does not obtain the highest ROC-AUC in this setting, it achieves the best accuracy, macro precision, macro recall, and macro F1-score. This suggests that the hybrid representation improves the final class decisions and class-wise balance, even if the probability ranking measured by ROC-AUC is slightly better for the Classical Matched model.

Although the Classical Matched model improves over the classical baseline, suggesting that an additional classical transformation stage is beneficial, the HQNN model still achieves the strongest overall performance. This indicates that the observed improvements are not solely due to additional parameters, but also to the expressive power of the quantum feature transformation.

\subsubsection{Class-Wise Performance}
\begin{figure}[t!]
    \centering
    \includegraphics[width=.85\linewidth]{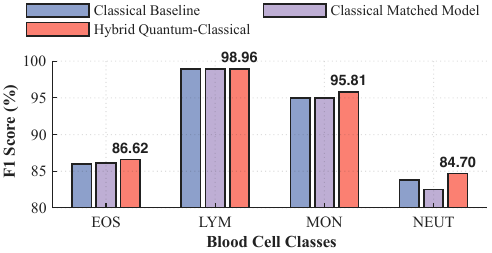}
    \vspace{-5pt}
    \caption{Per-class F1-score comparison for the evaluated models on the Blood Cell Images Dataset.}
    \label{fig:4class_eachclass_f1}
\end{figure}

\begin{figure}[t!]
    \centering
    \includegraphics[width=.8\linewidth]{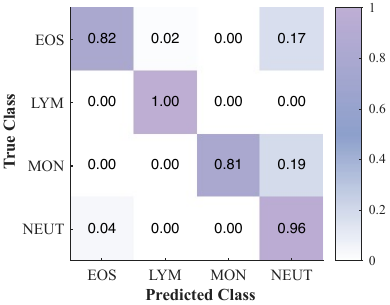}
    \vspace{-5pt}
    \caption{Confusion matrix of the HQNN on the Blood Cell Images Dataset.}
    \label{fig:4class_confusion}
\end{figure}

To better understand model behavior, a class-wise analysis using F1-scores for each blood cell category is reported in Fig.~\ref{fig:4class_eachclass_f1} and the confusion matrix of True vs. Predicted Class is reported in Fig.~\ref{fig:4class_confusion}.

Overall, all models achieved strong performance across most classes, with particularly high scores observed for lymphocytes, which consistently achieved near-perfect classification performance (F1-score close to 1.0) across all models. This indicates that lymphocytes are well-separated in the feature space and relatively easy to classify.

In contrast, monocytes and neutrophils represent more challenging classes. The baseline classical model achieved lower F1-scores for these categories (e.g., neutrophils around 0.79–0.83), indicating higher confusion between similar cell types. This is expected due to morphological similarities and overlapping visual features.

The HQNN model demonstrates noticeable improvements in these more challenging classes. In particular, the F1-score for neutrophils increases to approximately 0.8470, while improvements are also observed for monocytes. These gains suggest that the quantum transformation enhances feature discrimination in regions of the feature space where class boundaries are less distinct.

Eosinophils also achieve consistently high performance across all models, with only minor variations, indicating that this class is relatively easy to distinguish.

The confusion matrix analysis further supports these findings, showing that most misclassifications occur between monocytes and neutrophils, while other classes exhibit minimal confusion. The HQNN model reduces these misclassifications, leading to improved overall robustness.

In summary, while all models perform well on clearly separable classes, the HQNN approach provides the most significant benefits in challenging classification scenarios, improving class-wise balance and reducing confusion between visually similar cell types.
\subsection{PBC Dataset Results}

\subsubsection{Convergence Analysis}

\begin{figure}[t!]
    \centering
    \includegraphics[width=\linewidth]{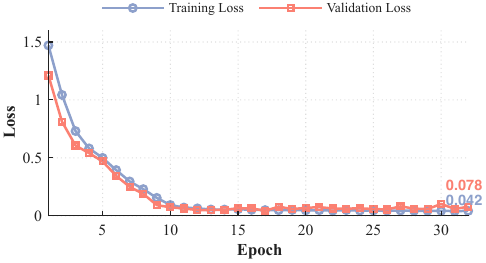}
    \vspace{-15pt}
    \caption{Training and validation loss curves of the HQNN model on the PBC dataset.}
    \label{fig:8class_training_loss}
\end{figure}

Fig.~\ref{fig:8class_training_loss} shows the training and validation loss curves of the HQNN model on the PBC Dataset, respectively. The loss decreases sharply during the initial epochs and stabilizes at a very low value afterward, indicating stable convergence and effective optimization. Despite the increased difficulty of the 8-class classification task, no major oscillations or divergence behavior are observed.

\begin{figure}[t!]
    \centering
    \includegraphics[width=\linewidth]{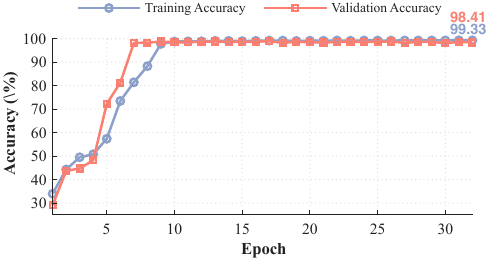}
    \vspace{-15pt}
    \caption{Training and validation accuracy curves of the HQNN model on the PBC dataset.}
    \label{fig:8class_training_accuracy}
\end{figure}

Fig.~\ref{fig:8class_training_accuracy} shows that the HQNN model reaches very high validation accuracy after only a few epochs and remains stable throughout training. The close alignment between training and validation curves suggests good generalization and no evident overfitting. This behavior is consistent with the near-saturated but still slightly superior classification performance of the HQNN model on this dataset.

\subsubsection{Overall Performance}
The performance of the evaluated models on the PBC Dataset, in terms of overall accuracy, macro-averaged precision, recall, and F1-score, together with macro and weighted ROC-AUC, is summarized in Table~\ref{tab:8class_overall_results}.

\begin{table}[t!]
\centering
\caption{Overall performance comparison on the PBC dataset}
\label{tab:8class_overall_results}
\resizebox{\columnwidth}{!}{%
\begin{tabular}{lccc}
\toprule
Metric & Classical Matched & Classical Baseline & HQNN \\
\midrule
Accuracy         & 0.9958 & 0.9957 & \textbf{0.9967} \\
Macro Precision  & 0.9876 & 0.9855 & \textbf{0.9857} \\
Macro Recall     & 0.9834 & 0.9849 & \textbf{0.9882} \\
Macro F1         & 0.9854 & 0.9851 & \textbf{0.9869} \\
ROC-AUC Macro    & 0.9977 & 0.9976 & \textbf{0.9955} \\
ROC-AUC Weighted & 0.9980 & 0.9971 & \textbf{0.9942} \\
\bottomrule
\end{tabular}%
}
\end{table}

\begin{figure}[t!]
    \centering
    \includegraphics[width=\linewidth]{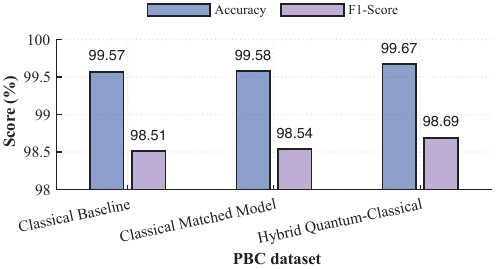}
    \vspace{-15pt}
    \caption{Comparison of overall accuracy and macro F1-score for the evaluated models on the PBC dataset.}
    \label{fig:8class_f1_acc}
\end{figure}

As shown in Fig.~\ref{fig:8class_f1_acc}, all models achieved very high performance, indicating that the extracted feature representations are highly discriminative. However, consistent differences are observed between the evaluated architectures.

The HQNN model achieved the highest macro F1-score of 0.9869, outperforming both the Classical Matched model (0.9854) and the classical baseline (0.9851). Although the absolute improvement is smaller compared to the 4-class scenario (approximately +0.1\%–0.2\%), this is expected given the near-saturation performance levels on this dataset.

Similarly, the HQNN model achieves the highest macro recall (0.9882) and maintains competitive macro precision (0.9857), indicating a well-balanced performance across all classes. In terms of overall accuracy, the HQNN model also achieves the best result (0.9967), slightly improving upon the Classical Matched model (0.9958) and the classical baseline (0.9957).

The ROC-AUC values are consistently high across all models (close to 1.0), reflecting excellent separability between classes. The HQNN model achieves competitive ROC-AUC performance, although the highest ROC-AUC values in this setting are obtained by the Classical Matched model.

Although the performance gap between models is smaller in this setting, the HQNN model consistently achieves the best or near-best results across all evaluation metrics. This suggests that while classical models already perform extremely well, the quantum-enhanced representation still provides measurable gains, particularly in maintaining balanced performance across all classes.

\subsubsection{Class-Wise Performance}

\begin{figure}[t!]
    \centering
    \includegraphics[width=\linewidth]{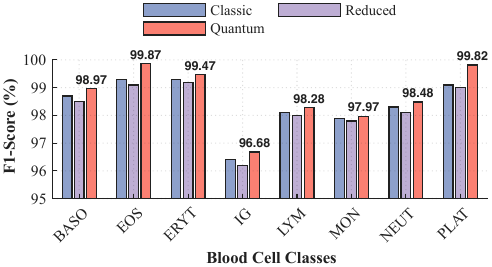}
    \vspace{-15pt}
    \caption{Per-class F1-score comparison for the evaluated models on the PBC dataset.}
    \label{fig:8class_eachclass_f1}
\end{figure}

A detailed class-wise analysis to better understand the behavior of the models across the eight blood cell categories is reported in Fig.~\ref{fig:8class_eachclass_f1}.

Overall, most classes achieve near-perfect performance across all models. In particular, eosinophils and platelets consistently achieve F1-scores close to 1.0, indicating that these classes are highly separable and easily distinguishable. Similarly, erythrocytes and basophils also achieve very high F1-scores, demonstrating strong classification performance.

More challenging classes include immature granulocytes (IG), monocytes, and neutrophils, where slightly lower F1-scores are observed. For example, the classical models achieve F1-scores around 0.95–0.97 for IG and slightly lower values for neutrophils. These classes present greater intra-class variability and inter-class similarity, making them hardly distinguishable.

The Hybrid Quantum-Classical model demonstrates improvements in these challenging classes. In particular, the F1-score for neutrophils increases to approximately 0.9848, while improvements are also observed for monocytes and IG cells. This indicates that the quantum-enhanced feature transformation is especially beneficial in resolving subtle differences between visually similar cell types.
Lymphocytes also achieve consistently high performance across all models, although slight variations are observed depending on the architecture. The HQNN model maintains competitive performance while improving balance across all classes.

\begin{figure}[t!]
    \centering
    \includegraphics[width=.9\linewidth]{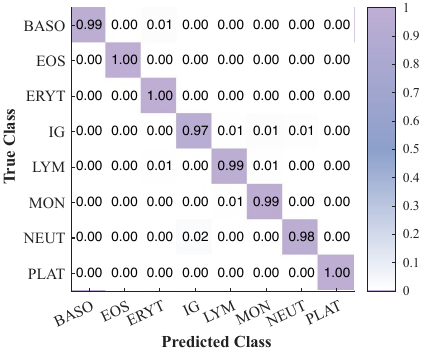}
    \vspace{-5pt}
    \caption{Confusion matrix of the HQNN model on the PBC dataset.}
    \label{fig:8class_confusion}
\end{figure}

The confusion matrix analysis, shown in Fig.~\ref{fig:8class_confusion}, further confirms that most misclassifications occur between biologically similar classes, particularly among granulocyte-related categories. The HQNN model reduces these misclassifications, leading to more consistent predictions across all classes.

In summary, while all models achieve near-saturated performance on easily distinguishable classes, the HQNN approach provides measurable improvements in more challenging categories, enhancing class-wise balance and robustness in fine-grained classification scenarios.

\subsection{Effect of Training Dataset Size}

\begin{table}[t!]
\centering
\caption{Accuracy and F1-score comparison using 25\% and 100\% of the dataset}
\label{tab:Dataset_Comparison}
\resizebox{\columnwidth}{!}{%
\begin{tabular}{llccc}
\toprule
Metric & Dataset Size & Classical Matched & Classical Baseline & HQNN \\
\midrule
Accuracy & 25\%  & 93.41\% & 93.06\% & \textbf{94.55\%} \\
Accuracy & 100\% & 94.29\% & 93.75\% & \textbf{95.72\%} \\
\midrule
F1-score & 25\%  & 87.14\% & 86.50\% & \textbf{89.33\%} \\
F1-score & 100\% & 88.62\% & 87.84\% & \textbf{91.53\%} \\
\bottomrule
\end{tabular}%
}
\end{table}

To evaluate the sensitivity of the proposed HQNN to smaller training datasets, a typical problem in medical imaging contexts, we further compared its performance against the classical matched model and the classical baseline using 25\% and 100\% of the Blood Cell Images Dataset. The results are shown in Table~\ref{tab:Dataset_Comparison}.

The results show that reducing the dataset size from 100\% to 25\% leads to a performance decrease for all models, as expected. However, the HQNN maintains the highest performance in both settings. With only 25\% of the dataset, the HQNN achieves 94.55\% accuracy and 89.33\% macro F1-score, outperforming the classical matched model by 1.14 percentage points in accuracy and 2.19 percentage points in F1-score. It also outperforms the classical baseline by 1.49 percentage points in accuracy and 2.83 percentage points in F1-score.

When trained with the full dataset, the HQNN reaches 95.72\% accuracy and 91.53\% macro F1-score, again achieving the best results among the three models. Although the HQNN shows a slightly larger absolute degradation when moving from 100\% to 25\% of the dataset, it consistently preserves its advantage over both classical alternatives. This indicates that the proposed HQNN architecture remains effective under reduced data availability and that the quantum layer contributes positively not only in the full-data setting but also when the training set is substantially reduced.

\subsection{Comparison with Prior Work}

Table~\ref{tab:prior_work_comparison} contextualizes our architectures against a cross-section of recent blood cell classification literature. We include representative studies on public blood cell classification benchmarks, prioritizing works that report overall accuracy and, when available, macro F1-score.

\begin{table}[h!]
\centering
\caption{Comparison with representative prior work on blood cell classification.}
\label{tab:prior_work_comparison}
\resizebox{\columnwidth}{!}{%
\begin{tabular}{llccc}
\toprule
Work & Dataset / Task & Classes & Accuracy & Macro F1 \\
\midrule
Acevedo et al.~\cite{acevedo2019peripheral} & PBC dataset & 8 & 96.2\% & -- \\
Girdhar et al.~\cite{girdhar2022wbc} & Blood Cell Images & 4 & 98.55\% & -- \\
Bayat et al.~\cite{bayat2022multiattention}& Blood Cell Images (Shenggan) & 4 & 99.69\% & -- \\
Islam et al.~\cite{islam2024explainable}& Blood Cell Images & 4 & 99.12\% & 99.0\% \\
Karaddi et al.~\cite{karaddi2026custom}& Blood Cell Images & 4 & 99.64\% & 99.34\% \\
\midrule
Ours (Classical Matched) & Blood Cell Images & 4 & 94.29\% & 88.62\% \\
Ours (Classical Baseline) & Blood Cell Images & 4 & 93.75\% & 87.84\% \\
Ours (HQNN) & Blood Cell Images & 4 & 95.72\% & 91.53\% \\
\midrule
Ours (Classical Matched) & PBC dataset & 8 & 99.58\% & 98.54\% \\
Ours (Classical Baseline) & PBC dataset & 8 & 99.57\% & 98.51\% \\
Ours (HQNN) & PBC dataset & 8 & 99.67\% & 98.69\% \\
\bottomrule
\end{tabular}
}
\end{table}

Representative prior work reports very strong results across several public blood cell classification benchmarks. For example, Acevedo et al. report 96.2\% accuracy on the PBC dataset, while Girdhar et al., Bayat et al., Islam et al., and Karaddi et al. report accuracies above 98\% on related but not identical blood cell datasets and class taxonomies. These results confirm that modern CNN-based pipelines are already highly competitive, especially on well-curated public benchmarks.

Comparing these numbers directly requires careful nuance. The literature features a wide variety of datasets, preprocessing techniques, and class boundaries. Furthermore, many studies rely exclusively on overall accuracy, a metric that easily obscures poor performance on minority or visually ambiguous classes. To ensure a rigorous evaluation of class-wise balance, our analysis relies heavily on macro-averaged metrics.

Against this established backdrop, our framework proves highly competitive. On the PBC dataset, the Classical Baseline achieves 99.57\% accuracy and 98.51\% macro F1-score, while the Classical Matched model reaches 99.58\% accuracy and 98.54\% macro F1-score. Integrating the quantum layer edges these metrics up to 99.67\% and 98.69\%, respectively. Because the classical models are already near the performance ceiling for this dataset, the absolute gains are naturally constrained, yet they remain remarkably consistent across all evaluated metrics.

The advantages of the HQNN architecture become much more obvious in the 4-class setting. Here, the HQNN's macro F1-score increased to 91.53\%, a nearly 3\% boost from 87.84\% in the classical baseline and 88.62\% in the Classical Matched model. This gain demonstrates that quantum-enhanced representations provide the most value in classification tasks that have not yet reached performance saturation and where subtle, fine-grained distinctions are critical.

Rather than simply chasing top-line accuracy, a primary goal of this study is isolating the actual impact of the quantum component. By directly comparing the quantum circuit to a functionally matched classical layer with comparable parameter overhead, we ensure that the observed improvements stem from the unique feature transformations of the quantum space, not merely from an influx of new trainable parameters. Ultimately, this controlled approach not only matches the performance of state-of-the-art classical pipelines but also establishes a clear, empirical baseline for how quantum representations function within medical imaging.

\subsection{Hardware Results}
\begin{table}[t!]
\centering
\caption{Performance on IBM quantum hardware (Blood Cell Images Dataset).}
\label{tab:hardware_results}
\footnotesize
\begin{tabular}{lcccccc}
\toprule
Accuracy & Macro F1-score & EOS & LYM & MON & NEUT \\
\midrule
 0.9000 & 0.8963 & 0.9091 & 0.9524 & 0.8235 & 0.9000 \\
\bottomrule
\end{tabular}
\end{table}

To further validate the practical applicability of the proposed HQNN model, experiments were conducted on real quantum hardware. Due to hardware constraints and execution time limitations, inference was performed on a reduced and balanced subset of the Blood Cell Images Dataset, consisting of 40 samples (10 per class).

The hardware experiment was executed on the IBM Quantum backend \texttt{ibm\_fez}, accessed through the Qiskit Runtime service and integrated with PennyLane using the \texttt{qiskit.remote} device.

The quantitative results are summarized in Table~\ref{tab:hardware_results}.

For comparison purposes, the same subset was first evaluated using the quantum simulator. The simulator achieved an accuracy of 0.9250 and a macro F1-score of 0.9246, providing a reference for ideal (noise-free) quantum execution.

When executing the model on real quantum hardware, a slight performance degradation was observed. The hardware execution achieved an evaluation accuracy of 0.9000 on the hardware subset and a macro F1-score of 0.8963, indicating a performance drop of approximately 2.5\%–3\% compared to the simulator. This degradation is expected due to the presence of quantum noise, limited qubit coherence times, and sampling errors inherent to current Noisy Intermediate-Scale Quantum (NISQ) devices~\cite{RevModPhys.94.015004, VQAlgorithms}.

A class-wise analysis reveals that the model maintains strong performance across most categories, particularly for lymphocytes, which achieved an F1-score of 0.9524, confirming that well-separated classes remain robust even under noisy quantum execution. Eosinophils and neutrophils also show high performance, with F1-scores of approximately 0.9091 and 0.9000, respectively.
However, a more noticeable performance drop is observed for monocytes, where the F1-score decreases to 0.8235. This aligns with previous observations in the classical and hybrid experiments, where monocytes represent a more challenging class due to higher intra-class variability and visual similarity with other cell types.

Overall, the results demonstrate that the proposed HQNN model remains robust when deployed on real quantum hardware, with only a moderate degradation in performance. This highlights the feasibility of applying quantum-enhanced machine learning models in practical scenarios, even with the current limitations of NISQ devices.

\subsection{Discussion}

The experimental results across both the Blood Cell Images Dataset and PBC datasets provide several important insights into the effectiveness of the proposed HQNN architecture.

First, the HQNN model consistently achieves the best accuracy, macro recall, and macro F1-score across both datasets, while remaining competitive on macro precision and ROC-AUC. The improvements are more pronounced in the Blood Cell Images Dataset setting, where gains of up to 3\% in macro F1-score are observed, indicating that the quantum feature transformation contributes to more discriminative representations. In the more challenging 8-class scenario, where all models achieve near-saturated performance, the HQNN model still provides consistent, albeit smaller, improvements. This behavior suggests that the benefits of quantum-enhanced representations are particularly relevant in moderately complex classification tasks and remain stable even as performance approaches upper limits.

A key observation is that the HQNN model demonstrates the most significant improvements in challenging classes, such as monocytes and neutrophils, which exhibit higher inter-class similarity. This indicates that the quantum transformation enhances feature separability in regions of the feature space where classical models struggle. At the same time, performance on well-separated classes remains consistently high across all models, confirming that the HQNN model does not degrade performance in simpler cases.

The comparison with the classical baseline and the Classical Matched model also highlights an important aspect: adding an extra classical transformation stage yields a modest improvement, but the HQNN model still achieves superior or more balanced results. This suggests that the observed improvements are not solely due to parameter count, but also to the different transformation properties introduced by the quantum layer.

The evaluation on real quantum hardware further demonstrates the practical feasibility of the proposed approach. Although a moderate performance degradation is observed compared to the simulator, the model maintains strong accuracy and F1-score, confirming its robustness under realistic noisy conditions. This result is particularly relevant, as it validates the applicability of HQNN models beyond idealized simulations.

Despite these promising results, some limitations remain and motivate future work. First, the evaluation on quantum hardware was performed on a reduced subset of the dataset due to current hardware constraints. Scaling the approach to larger datasets and full-batch inference remains a challenge. Second, while the HQNN model shows improvements, the gains become marginal in near-saturated scenarios, suggesting that future work should investigate more complex datasets or tasks where quantum advantages may be more pronounced.

Additionally, the current study focuses on a specific integration strategy between classical and quantum components. Future research could explore alternative encoding schemes, circuit topologies, and HQNN training strategies to better exploit quantum representations. Finally, advancements in quantum hardware, particularly in reducing noise and increasing qubit fidelity, are expected to further enhance the practical impact of HQNN models.

In summary, the results demonstrate that HQNN models provide consistent and meaningful improvements over classical approaches, particularly in challenging classification scenarios, while remaining robust under real hardware constraints.

\section{Conclusion}

By enforcing a strictly controlled evaluation framework, this study clarifies the specific diagnostic value of HQNNs in blood cell classification. Rather than comparing disparate pipelines, we held the feature extraction backbone, bottleneck, classifier head, and training conditions constant across three distinct architectures: a baseline classical model, a capacity-matched classical network, and the proposed HQNN pipeline. This rigid design successfully isolated the quantum layer's true contribution to the classification process.

Across both public datasets, the HQNN model consistently achieved the strongest performance on the main class-balanced metrics, particularly macro recall and macro F1-score. The advantages are most striking in the Blood Cell Images Dataset scenario, where the HQNN architecture markedly improved macro F1-scores and successfully resolved visual ambiguities between notoriously difficult classes, such as monocytes and neutrophils. Even in the highly saturated 8-class task, where classical models are already pushed to their limits, the quantum model secured consistent, although incremental, gains in both accuracy and class-wise balance.

Crucially, benchmarking against the functionally matched Classical Matched model supports the interpretation that these improvements stem from the distinct transformation induced by the quantum layer, rather than merely from adding another intermediate classical stage. Furthermore, testing on physical quantum hardware confirms that the proposed architecture retains its discriminative power and remains robust, even when subjected to the inherent noise of current quantum devices.

Ultimately, these findings validate the tangible benefits of quantum feature transformations for resolving complex medical image classifications. By anchoring the analysis in a fair, reproducible baseline, this research establishes a firm empirical foundation for integrating quantum-enhanced learning into the next generation of clinical diagnostic tools.

\section*{Acknowledgment}
This work is partially funded by national funds through FCT – Fundação para a Ciência e a Tecnologia, I.P., and, when eligible, co-funded by EU funds under project/support UID/50008/2025 – Instituto de Telecomunicações, with DOI identifier - https://doi.org/10.54499/UID/50008/2025. This work was supported in part by the NYUAD Center for Quantum and Topological Systems (CQTS), funded by Tamkeen under the NYUAD Research Institute grant CG008.

\end{spacing}

\bibliographystyle{IEEEtran}

\bibliography{refs}

\end{document}